\documentclass{article}

\usepackage{microtype}
\usepackage{graphicx}
\usepackage{subcaption}
\usepackage{booktabs} 
\usepackage{tabularx}
\usepackage{hyperref}
\usepackage{xcolor}
\usepackage{enumitem}
\usepackage{bbm}
\usepackage{tcolorbox}
\usepackage{threeparttable}
\definecolor{lightbluebg}{RGB}{220, 230, 255} 
\definecolor{darkblueborder}{RGB}{0, 0, 190}

\usepackage[preprint]{icml2026}

\usepackage{amsmath}
\usepackage{amssymb}
\usepackage{mathtools}
\usepackage{amsthm}

\usepackage[capitalize,noabbrev]{cleveref}
\usepackage[textsize=tiny]{todonotes}
\usepackage{multirow}
\usepackage{tikz}
\usepackage{makecell}

\newcommand*\levelE{\tikz[baseline=(char.base)]{\node[shape=circle,draw=black,fill=green!60,inner sep=1pt] (char) {\tiny\sffamily E};}}
\newcommand*\levelH{\tikz[baseline=(char.base)]{\node[shape=circle,draw=black,fill=orange!70,inner sep=1pt] (char) {\tiny\sffamily H};}}
\newcommand*\levelC{\tikz[baseline=(char.base)]{\node[shape=circle,draw=black,fill=cyan!60,inner sep=1pt] (char) {\tiny\sffamily C};}}
\newcommand*\levelU{\tikz[baseline=(char.base)]{\node[shape=circle,draw=black,fill=purple!60,inner sep=1pt] (char) {\tiny\sffamily U};}}
\newcommand*\levelO{\tikz[baseline=(char.base)]{\node[shape=circle,draw=black,fill=orange!80,inner sep=1pt] (char) {\tiny\sffamily O};}}
\theoremstyle{plain}

\theoremstyle{definition}

\theoremstyle{remark}

\icmltitlerunning{The \texttt{CompMath-MCQ} Dataset: Are LLMs Ready for Higher-Level Math?}

\begin{document}

\twocolumn[
  \icmltitle{The \texttt{CompMath-MCQ} Dataset: Are LLMs Ready for Higher-Level Math?}

  \icmlsetsymbol{equal}{*}
  \begin{icmlauthorlist}
    \icmlauthor{Bianca Raimondi}{equal,disi}
    \icmlauthor{Francesco Pivi}{equal,disi}
    \icmlauthor{Davide Evangelista}{disi}
    \icmlauthor{Maurizio Gabbrielli}{disi}
  \end{icmlauthorlist}

  \icmlaffiliation{disi}{Department of Computer Science and Engineering, University of Bologna, Bologna, 40126, Italy}
  \icmlcorrespondingauthor{Bianca Raimondi}{bianca.raimondi3@unibo.it}

  \icmlkeywords{Machine Learning, ICML}

  \vskip 0.3in
]
\printAffiliationsAndNotice{\icmlEqualContribution}

\begin{abstract}
The evaluation of Large Language Models (LLMs) on mathematical reasoning has largely focused on elementary problems, competition-style questions, or formal theorem proving, leaving graduate-level and computational mathematics relatively underexplored. We introduce \texttt{\textbf{CompMath-MCQ}}, a new benchmark dataset for assessing LLMs on advanced mathematical reasoning in a multiple-choice setting. The dataset consists of 1{,}500 originally authored questions by professors of graduate-level courses, covering topics including Linear Algebra, Numerical Optimization, Vector Calculus, Probability, and Python-based scientific computing. Three option choices are provided for each question, with exactly one of them being correct. \\
To ensure the absence of data leakage, all questions are newly created and not sourced from existing materials. The validity of questions is verified through a procedure based on cross-LLM disagreement, followed by manual expert review. By adopting a multiple-choice format, our dataset enables objective, reproducible, and bias-free evaluation through \texttt{lm\_eval} library. Baseline results with state-of-the-art LLMs indicate that advanced computational mathematical reasoning remains a significant challenge. We release \texttt{\textbf{CompMath-MCQ}} at the following link:
\url{https://github.com/biancaraimondi/CompMath-MCQ.git}
\end{abstract}

\section{Introduction}
\label{sec:intro}

Large Language Models (LLMs) have recently demonstrated strong performance on a variety of mathematical reasoning tasks, motivating the development of increasingly challenging benchmarks to assess their capabilities. Early datasets such as GSM8K~\cite{cobbe2021gsm8k} and MATH~\cite{hendrycks2021measuring} have played a central role in this progress, providing standardized evaluations for arithmetic reasoning and competition-style problem solving. More recent efforts have introduced harder benchmarks inspired by mathematical olympiads \cite{gao2024omni,he2024olympiadbench,zheng2021minif2f} or formal theorem proving \cite{chen2023theoremqa}, aiming to probe the limits of current models.

Despite this rapid evolution, the evaluation of LLMs on \emph{advanced, graduate-level mathematics} remains relatively underexplored. Existing benchmarks predominantly focus on one of three settings: \textit{(i)} elementary to undergraduate-level problem solving; \textit{(ii)} olympiad-style questions emphasizing ingenuity and short solutions; or \textit{(iii)} formal theorem proving, which requires symbolic representations and specialized tooling. As a consequence, current evaluations provide limited insight into whether LLMs possess the conceptual, technical, and computational understanding expected at the graduate and PhD levels, particularly in applied and computational domains such as Linear Algebra, Optimization, Probability, and scientific computing.

Meanwhile, the multiple-choice question (MCQ) format, which is commonly used in graduate coursework, qualifying exams and structured assessments, has received comparatively little attention in advanced mathematical benchmarking. While MCQs are sometimes perceived as less expressive than free-response problems, they offer several advantages that are particularly relevant for large-scale evaluation of LLMs: objective and unambiguous scoring, reproducibility across inference settings, and the ability to probe fine-grained conceptual distinctions through carefully designed distractors. These properties are especially important when comparing models with different architectures, sizes, or prompting strategies.

In this work, we introduce \texttt{CompMath-MCQ}, a new benchmark dataset of $1{,}500$ MCQs designed to address gaps in the literature and to evaluate LLMs on \emph{graduate- and PhD-level computational mathematics}. The dataset is constructed from educational material commonly taught in master-level university courses and spans five core areas: Linear Algebra, Numerical Optimization, Vector Calculus, Probability, and Python programming for scientific computing. All questions were written manually by the authors, who are professors of these subjects. Each question consists of three possible answers, only one of which is correct, and is designed to assess conceptual understanding, theoretical insight, or technical reasoning rather than rote computation or formal proof writing.

A key distinguishing feature of \texttt{CompMath-MCQ} is that \emph{none of the questions were sourced from existing textbooks, online repositories, or prior benchmarks}. All items were originally created by the authors specifically for this dataset. As a result, our dataset is inherently free from data leakage with respect to the training corpora of contemporary LLMs and therefore provides a more faithful assessment of their generalization capabilities. This contrasts with many existing benchmarks, where partial overlap with publicly available educational content or online problem collections can inadvertently inflate measured performance.

To validate the consistency of created questions, we adopt a two-stage validation procedure (see Section~\ref{sec:compmath-mcq} for details). First, all questions are evaluated by collecting multiple state-of-the-art LLMs' answers. Then, we run statistical tests on the per-question error rate to identify questions on which multiple models produce incorrect answers and manually review them to verify that the problem statements are unambiguous and that the designated correct answers are indeed correct. This process allows us to systematically detect and eliminate potential errors or ambiguities while avoiding reliance on any single model as an oracle.

Beyond validation, \texttt{CompMath-MCQ} also addresses a common source of bias in mathematical benchmarking: the interpretation of open-ended answers. In free-response settings, evaluation often relies on heuristic matching of model outputs~\cite{cobbe2021gsm8k}, symbolic simplification and equivalence checking~\cite{hendrycks2021measuring}, or subjective human judgment~\cite{lewkowycz2022solving}, all of which can introduce inconsistencies and model-dependent biases. By adopting a multiple-choice format with a single correct answer, \texttt{CompMath-MCQ} enables fully automatic and deterministic evaluation, ensuring fair and reproducible comparison across models exploiting the \texttt{lm\_eval} library~\cite{eval-harness} (see Section~\ref{sec:experiments} for evaluation details).

We position our dataset as a missing link in the current evaluation landscape: a standardized, leakage-free benchmark for assessing LLM performance on advanced computational mathematical reasoning in a multiple-choice setting, closely aligned with graduate-level academic curricula. By releasing \texttt{CompMath-MCQ}, we aim to support more rigorous evaluation of LLMs on high-level mathematics and to provide a reliable testing ground for future advances in mathematical reasoning beyond the undergraduate level.

The contributions of this work are as follows: \textit{(i)} we introduce \texttt{CompMath-MCQ}, a benchmark dataset of $1{,}500$ manually authored MCQs for evaluating LLMs on graduate- and PhD-level computational mathematics, covering topics such as Linear Algebra, Optimization, Vector Calculus, and Probability, ensuring curriculum alignment and the absence of data leakage; \textit{(ii)} we propose a two-stage dataset validation framework combining automated model-agreement statistics (error rate, wrong-answer consensus, and statistical anomaly detection) with manual expert review to identify and correct ambiguous or mislabeled items, improving the reliability and internal consistency of the benchmark; and \textit{(iii)} we provide baseline evaluations with state-of-the-art LLMs, demonstrating that advanced computational mathematical reasoning remains a challenging setting and establishing reference performance levels for future work.

The work is organized as follows: in Section~\ref{sec:related_work}, we position our work in relation to the literature; in Section~\ref{sec:compmath-mcq}, we present our dataset and the validation framework; in Section~\ref{sec:experiments}, we explain the evaluation used to test the performance of LLMs; and in Sections~\ref{sec:discussion} and \ref{sec:conclusion}, we discuss the work and draw conclusions.

\section{Related Work}\label{sec:related_work}

\begin{table*}[ht]
\centering

\label{tab:comparison}
\small
\begin{tabular}{lccccc}
\toprule
Dataset & \# Problems & Level & Format & Data Leakage & Curriculum Aligned \\
\midrule
GSM8K \cite{cobbe2021gsm8k} & 8.5K & \levelE & Open-ended & High & No \\
MATH \cite{hendrycks2021measuring} & 12.5K & \levelH & Open-ended & High & No \\
HARP \cite{yue2024harp} & 5.4K & \levelO & Short answer & Medium & No \\
OmniMath \cite{gao2024omni} & 4.4K & \levelO & Open-ended & Medium & No \\
OlymMATH \cite{sun2025challenging} & 200 & \levelO & Closed-form  & Medium & No \\
MiniF2F \cite{zheng2021minif2f} & 488 & \levelO & Formal proof & Low & No \\
ProofNet \cite{azerbayev2023proofnet} & 371 & \levelU & Formal proof & Low & Partial \\
Orca-Math \cite{mitra2024orca} & 200K & Synthetic & Q\&A & Low & No \\
AoPS-Instruct \cite{lu2024aops} & 600K+ & \levelO & Q\&A & High & No \\
u-Math \cite{yue2024umath} & 1.1K & \levelU & Open-ended & Low & Partial \\
Uni-MMMu\cite{zhu2024mmumath} & 885 & \levelU & Open-ended & Low & Partial \\
MathVista \cite{lu2023mathvista} & $\sim$6K & Mixed & Mixed MCQ & Medium & No \\
WE-Math \cite{wang2024wmath} & 1.7K & \levelU & MCQ & Medium & No \\
Hard-Math \cite{fan2024hardmath} & 1.4K & \levelU  & Open-ended & Low & Yes \\
StatEval \cite{lu2025stateval} & 13.8K & \levelC \levelU & MCQ & High & Yes \\
\midrule
\textbf{CompMath-MCQ (Ours)} & \textbf{1.5K} & \levelU & \textbf{MCQ} & \textbf{None} & \textbf{Yes} \\
\bottomrule
\end{tabular}
\caption{Comparison of math reasoning benchmarks. Level markers: \levelE{} Elementary, \levelH{} High School, \levelC{} College/Undergrad, \levelU{} University/Graduate, \levelO{} Olympiads. Data leakage indicates likelihood of overlap with LLM training data: None (originally authored, first time available), Low (problems not published online and accessible on restriction), Medium (problems from recent competitions, with a portion of sources being public), High (Problems sourced from publicly available textbooks, websites, or online forums).}
\end{table*}

The evaluation of language models' mathematical reasoning abilities has evolved significantly over the last few years, targeting diverse skills while highlighting gaps that motivate our work.

Early benchmarks focused on elementary and undergraduate-level problems~\cite{sobhani2025mathmist,hendrycks2020measuring}. GSM8K~\cite{cobbe2021gsm8k} established a foundational dataset with $8.5$k human-curated elementary math word problems, while MATH~\cite{hendrycks2021measuring} introduced $12.5$k competition problems aimed at testing undergraduate knowledge. These datasets are approaching saturation, primarily testing logical reasoning over procedural mathematical execution. Moreover, their widespread online availability raises concerns about possible data contamination on newer models' training procedure.

Subsequent efforts shifted to competition and Olympiad-level mathematics to emphasize creative problem-solving. HRAP~\cite{yue2024harp} curated $5{,}409$ problems from U.S. national competitions (AMC, AIME, USAMO) with human-annotated difficulty labels. Similarly, OmniMath~\cite{gao2024omni} and OlymMATH~\cite{sun2025challenging} provide collections of $4{,}428$ and $200$ expert-annotated Olympiad problems, respectively. While these benchmarks effectively challenge frontier models, they prioritize non-standard, procedurally intensive problems.

Formal proof verification represents another research direction~\cite{yu2025formalmath}. MiniF2F~\cite{zheng2021minif2f} translates Olympiad problems into \emph{Lean} formal language, and ProofNet~\cite{azerbayev2023proofnet} targets university-level proofs by decomposing principal problems into sub-problems to guide theorem construction. These approaches demand specialized logical representations and tools, diverging from pure reasoning development.

Synthetic and large-scale datasets address data scarcity for everyday problems~\cite{tang2024mathscale}. Orca-Math~\cite{mitra2024orca} generates $200$k problems using GPT-4 Turbo agents, while AoPS-Instruct~\cite{lu2024aops} produces over $600$k question-answer pairs from Olympiad online resources. StatEval~\cite{lu2025stateval} generated $13.8$k statistic questions from textbook content. However, synthetic methods inevitably mirror training data distributions, limiting their ability to assess true generalization and raising data leakage concerns.

Recent works target university-level and applied mathematics. HardMath \cite{fan2024hardmath} focuses on advanced topics like asymptotic methods, while u-Math~\cite{yue2024umath} introduces $1,100$ unpublished university problems across six subjects ($20\%$ of which are multimodal, including tables, figures, and graphs) in an open-ended format, though this introduces subjectivity and evaluation inconsistencies.

Multimodal benchmarks incorporate visual reasoning. Uni-MMMu~\cite{zhu2024mmumath} offers $885$ university-level multiple-choice problems with images across disciplines, MathVista~\cite{lu2023mathvista} aggregates $\sim$$6$k examples from $28$ datasets, and WE-Math~\cite{wang2024wmath} provides $1.7$k graduate-level multimodal multiple-choice questions.

Despite this rich landscape, a significant gap persists: no benchmark specifically evaluates LLMs on graduate-level computational and applied mathematics problems in a reproducible format free of data leakage. Olympiad problems stress non-standard creativity over procedural competence, while open-ended university benchmarks require subjective LLM-based evaluation that introduces inconsistencies.

In this work, we aim to fill this gap by providing problems that are aligned with graduate curricula, sourced and authored by professors teaching master-level courses. The multiple-choice format enables deterministic and reproducible evaluation through the \texttt{lm\_eval} library, eliminating the ambiguities inherent in parsing open-ended responses. By ensuring that all questions are originally created for this dataset and not drawn from existing materials, we guarantee the absence of data leakage and provide a more faithful assessment of model generalization capabilities. Finally, our emphasis on computational topics like Vector Calculus, Linear Algebra, Optimization, and applied methods, addresses the procedural and technical competencies critical for research applications, distinguishing our benchmark from competition-style datasets that prioritize mathematical ingenuity over systematic problem-solving skills.

\section{Materials and Methods}\label{sec:compmath-mcq}

This section describes the methodology adopted for the construction and validation of the \texttt{CompMath-MCQ} dataset. The guiding principles of the process are mathematical correctness, alignment with graduate-level curricula, absence of data leakage, and suitability for objective and reproducible evaluation of LLMs.

\subsection{The \texttt{CompMath-MCQ} dataset}

\begin{figure}[ht]
\begin{center}
\centerline{\includegraphics[width=\linewidth]{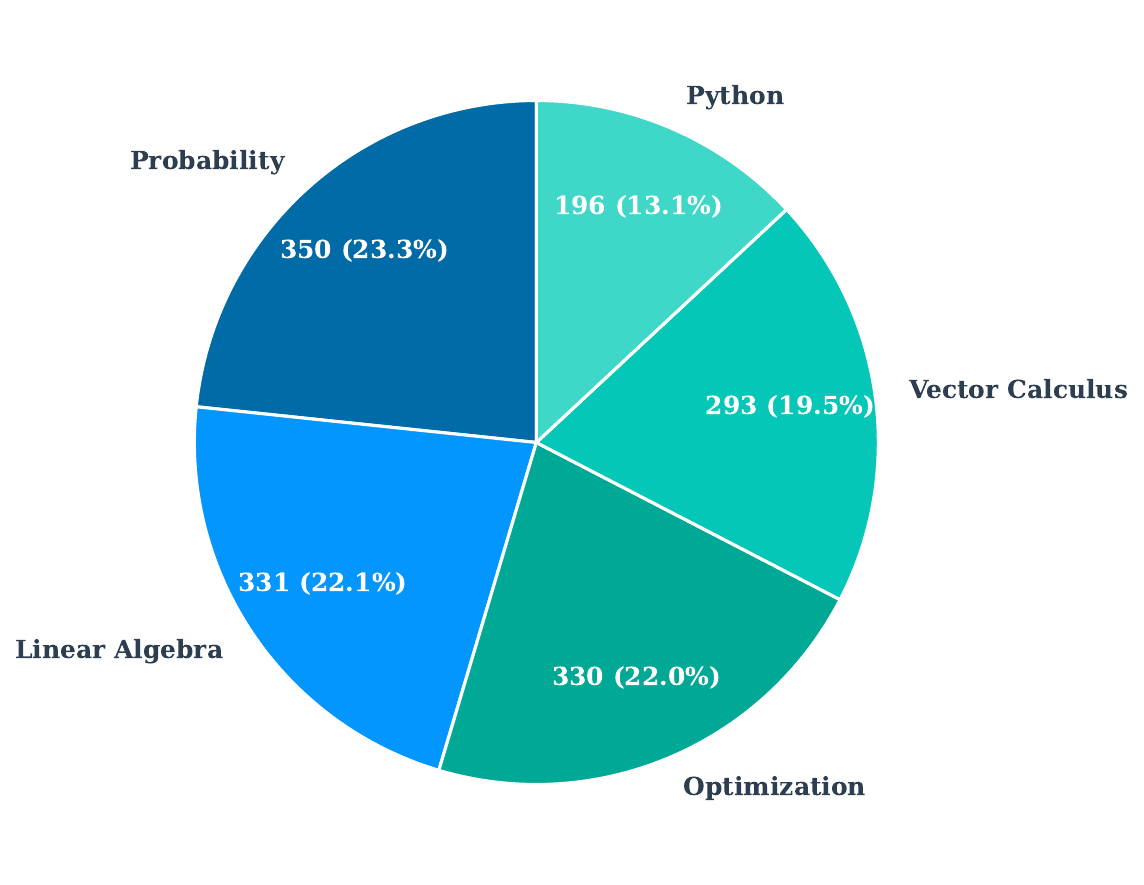}} 
\caption{Distribution of questions across mathematical topics in the CompMath-MCQ dataset. The dataset contains 1,500 questions spanning five core areas.}
\label{fig:pie_chart}
\end{center}
\end{figure}

The scope of the dataset is defined by the content of master-level courses in computational and applied mathematics. In particular, \texttt{CompMath-MCQ} covers Linear Algebra, Numerical Optimization, Vector Calculus, Probability, and Python-based scientific computing. Each question is associated with a primary topic, and the overall topic distribution is chosen to reflect the relative emphasis of these subjects in typical graduate curricula. In particular, topics such as Probability, Linear Algebra, and Optimization constitute the largest portion of the dataset, in line with their prominence in typical graduate curricula. Other topics, including Python programming and Vector Calculus, which are often taught alongside these subjects, are represented with a smaller number of questions; for example, Python accounts for only $13.1\%$ of the total dataset. The precise distribution of questions across categories, together with a summary of the main topics covered in each category, is reported in 
Figure~\ref{fig:pie_chart}. Collectively, these topics span the core material of a standard course in Numerical Linear Algebra or Computational Mathematics, ranging from matrix operations and Singular Value Decomposition (SVD) to gradient computations, Numerical Optimization methods, and fundamental concepts in probability. Questions involving Python contain both reasoning about algorithmic behaviour, numerical stability, or implementation choices, and syntactic details.

\subsection{Validation framework}
Each question is designed to assess conceptual understanding, theoretical insight, or technical reasoning, rather than rote computation or formal proof writing. Questions are formulated in a multiple-choice format with exactly three answer options, of which only one is correct. The remaining options are constructed as \emph{plausible distractors}, reflecting common misconceptions or typical reasoning errors observed in graduate-level education. As an example, consider the first question in Table \ref{tab:example_matrix}, which asks about the relationship between the singular values of a matrix $A \in \mathbb{R}^{m \times n}$ and the eigenvalues of $A^\top A$. While the correct answer is clearly \textbf{(c)}, students often choose either \textbf{(a)} or \textbf{(b)}, as these options are strongly associated with closely related concepts covered in the same area.

\begin{table*}[t]
\centering
\tcbox[
    colback=lightbluebg,     
    colframe=darkblueborder,   
    boxrule=1.2pt,             
    arc=3pt
]{
\small
\begin{tabularx}{\linewidth}{l X}
\toprule
Category & Representative Example (correct option in bold) \\
\midrule
\textbf{Linear Algebra} &
The singular values of $A$ are:
\begin{enumerate}[noitemsep, label=(\alph*)]
\item eigenvalues of $A^\top A$
\item absolute value of eigenvalues of $A^\top A$ 
\item \textbf{square roots of eigenvalues of $A^\top A$} (\checkmark)
\end{enumerate}\\
\midrule
\textbf{Optimization} &
A stationary point of $f(x,y) = x^2 - 4xy + 4y^2$ is:
\begin{enumerate}[noitemsep, label=(\alph*)]
\item \textbf{$\mathbf{(0,0)}$} (\checkmark)
\item $(2,2)$
\item no stationary points
\end{enumerate}\\
\midrule
\textbf{Vector Calculus} &
Let $f(x_1,x_2,x_3) = \sin x_1 \sin x_3 + x_2^2 - \cos x_1 \sin x_2$.
Then $\nabla f(0,0,\pi/2)$ equals:
\begin{enumerate}[noitemsep, label=(\alph*)]
\item $(1,0,1)$
\item \textbf{$\mathbf{(1,-1,0)}$} (\checkmark)
\item $(0,0,1)$
\end{enumerate}\\
\midrule
\textbf{Probability} &
Let $X$ be the face value of a fair six-sided die, rounded down to the nearest multiple of $2$ (with $X=0$ if the outcome is $1$).
The image of $X$ is:
\begin{enumerate}[noitemsep, label=(\alph*)]
\item $\{0,2,4\}$
\item \textbf{$\mathbf{\{0,2,4,6\}}$} (\checkmark)
\item $\{2,4,6\}$
\end{enumerate}\\
\midrule
\textbf{Python} &
Let \texttt{r = A[0]} be a one-dimensional array of shape \texttt{(3,)} obtained from a $3 \times 3$ matrix $A$.
What is the result of \texttt{r[1{:}3]}?
\begin{enumerate}[noitemsep, label=(\alph*)]
\item a two-dimensional array of shape $(1,2)$
\item \textbf{a one-dimensional array containing the last two elements of the first row} (\checkmark)
\item a copy of the entire row $r$
\end{enumerate}\\
\bottomrule
\end{tabularx}}
\caption{One representative example per category from the \texttt{CompMath-MCQ} dataset. Correct answers are highlighted for illustrative purposes only.}
\label{tab:example_matrix}
\end{table*}

Ensuring the correctness and internal consistency of the dataset is a central design requirement. To this end, we adopted a two-stage verification procedure combining automated screening with manual expert review.

In the first stage, we perform an automated consistency analysis based on model agreement statistics across a heterogeneous set of LLMs. The objective of this stage is not to assess model performance per se, but to identify questions whose labeled answer may be ambiguous, underspecified, or incorrect, as indicated by systematic disagreement across multiple independent models. 

Eight LLMs were selected for question validation: three closed-source models (\texttt{Gemini 3}, \texttt{GPT-5}, and \texttt{Claude Sonnet 4.5}) and five open-source models (\texttt{Llama-3.1 8B Instruct}, \texttt{Qwen2.5-Math 7B Instruct}, \texttt{Qwen3 4B Instruct}, \texttt{Qwen3 30B Instruct}, and \texttt{Qwen3-Coder 30B Instruct}). The closed-source models serve as a high-performance reference baseline, while the open-source models ensure transparency and reproducibility. Using multiple models with different development paradigms increases the robustness of the validation process and reduces the risk of model-specific bias, thereby strengthening the reliability of the results.

For each question, we record the answer predicted by each model and compute a set of per-question agreement indicators. 
Let $M$ denote the number of models and let $\hat{y}_i^{(m)}$ be the answer predicted by model $m$ for question $i$, with ground-truth label $y_i$. We define the \emph{per-question error rate} as the fraction of models predicting an incorrect answer:
\begin{align}\label{eq:error_rate}
\mathfrak{e}_i := \frac{1}{M} \sum_{m=1}^M \mathbbm{1}\left(\hat{y}_i^{(m)} \neq y_i\right),
\end{align}
where $\mathbbm{1}$ represents the indicator function, whose value is $1$ if the condition is verified and $0$ otherwise.
This quantity provides a coarse measure of question difficulty and highlights items that are frequently answered incorrectly across models.

\begin{figure}[t]
    \centering
    \resizebox{\linewidth}{!}{
    \begin{subfigure}[b]{0.49\linewidth}
        \centering
        \includegraphics[width=\textwidth]{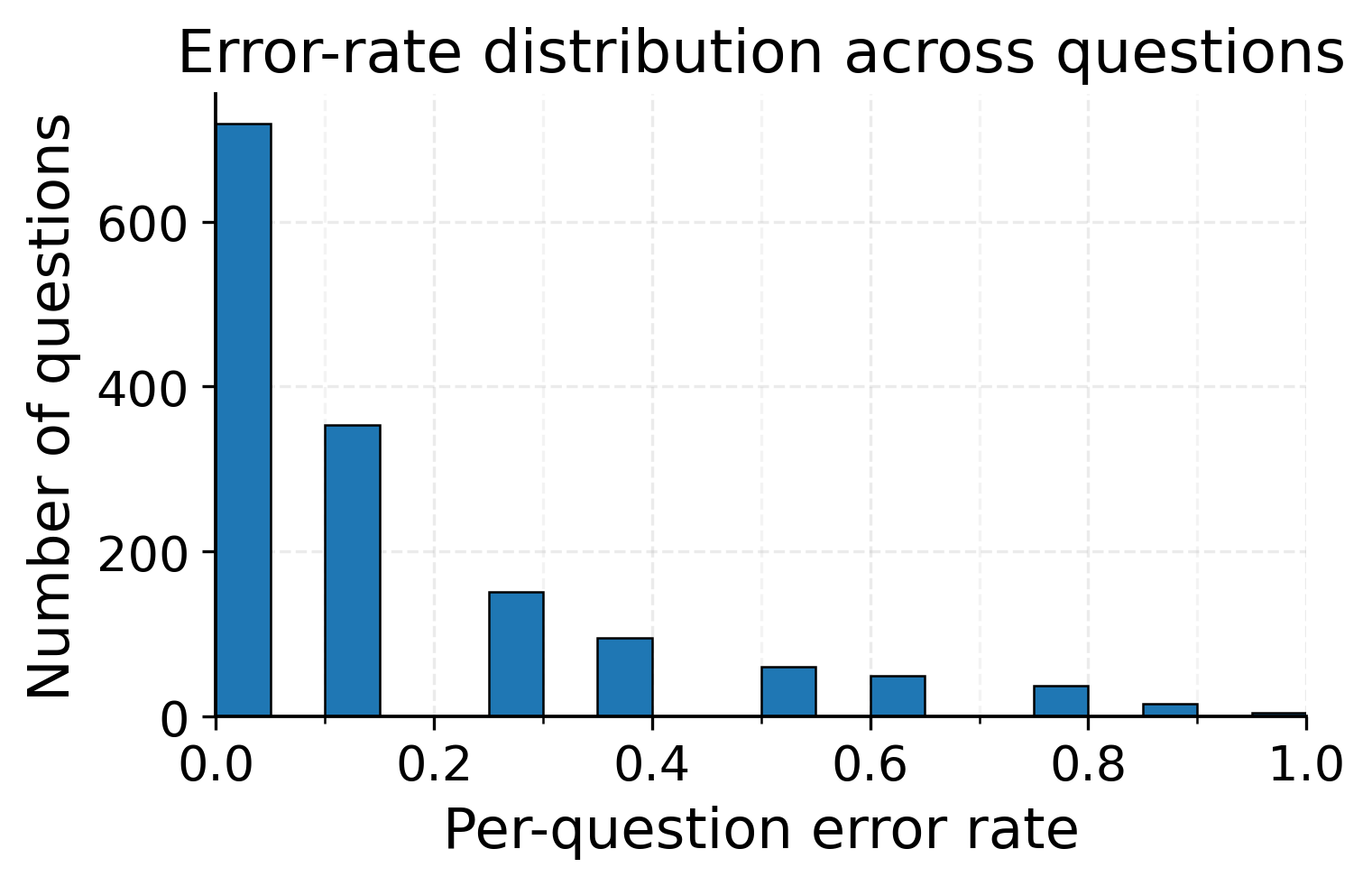}
        \caption{}
        \label{fig:error_rate_distribution}
    \end{subfigure}
    \hfill
    \begin{subfigure}[b]{0.49\linewidth}
        \centering
        \includegraphics[width=\textwidth]{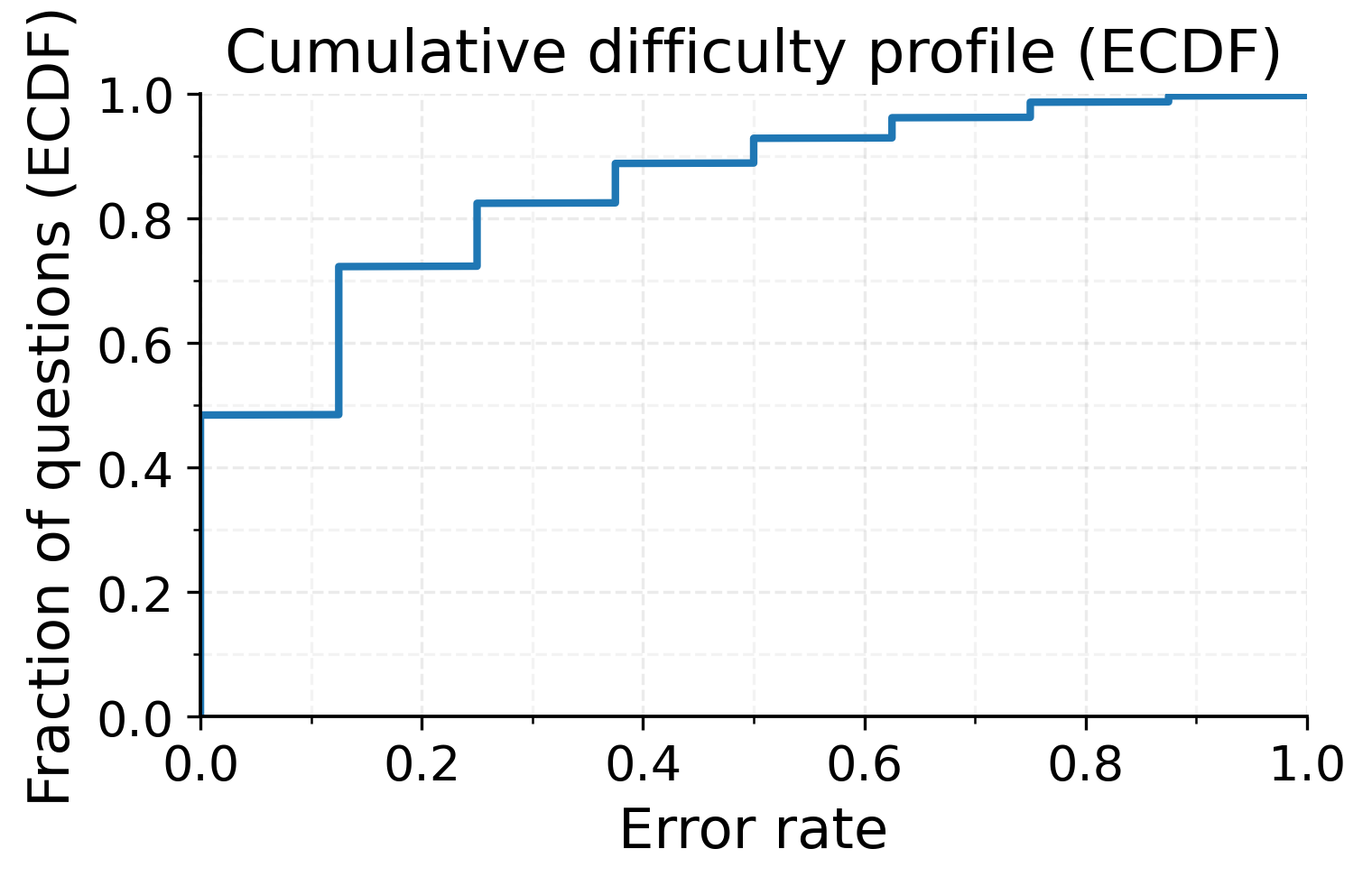}
        \caption{}
        \label{fig:error_rate_ecdf}
    \end{subfigure}}
    \caption{Distribution (a) and empirical cumulative distribution function (b) of the per-question error rate, defined as the fraction of models predicting an incorrect answer.}
    \label{fig:error_rate}
\end{figure}

Figure~\ref{fig:error_rate} reports the distribution and empirical cumulative distribution (ECDF) of the per-question error rate $\mathfrak{e}_i$ across the dataset. The strongly right-skewed distribution indicates that most questions are correctly answered by the majority of models, while a smaller subset exhibits high error rates, motivating targeted inspection.

However, the high error rate alone does not distinguish between genuinely hard questions and potentially flawed ones. To this end, we further analyze the structure of model disagreement by measuring the degree of \emph{wrong-answer consensus}, defined as the relative frequency of the most common incorrect option $\bar{y}_i \neq y_i$ among the models that answered incorrectly, i.e.
\begin{align}
    \mathfrak{c}_i := \frac{\max_{\bar{y}_i \neq y_i} \left\{\sum_{m=1}^M \mathbbm{1}\left(\hat{y}_i^{(m)} = \bar{y}_i\right) \right\}}{\sum_{m=1}^M \mathbbm{1}\left(\hat{y}_i^{(m)} \neq y_i\right)}.
\end{align}
High values of this measure indicate that models tend to converge to the same incorrect answer, suggesting the presence of a misleading distractor or an ambiguity in the question formulation.

\begin{figure}[t]
    \centering
    \resizebox{\linewidth}{!}{
    \begin{subfigure}[b]{0.49\linewidth}
        \centering
        \includegraphics[width=\textwidth]{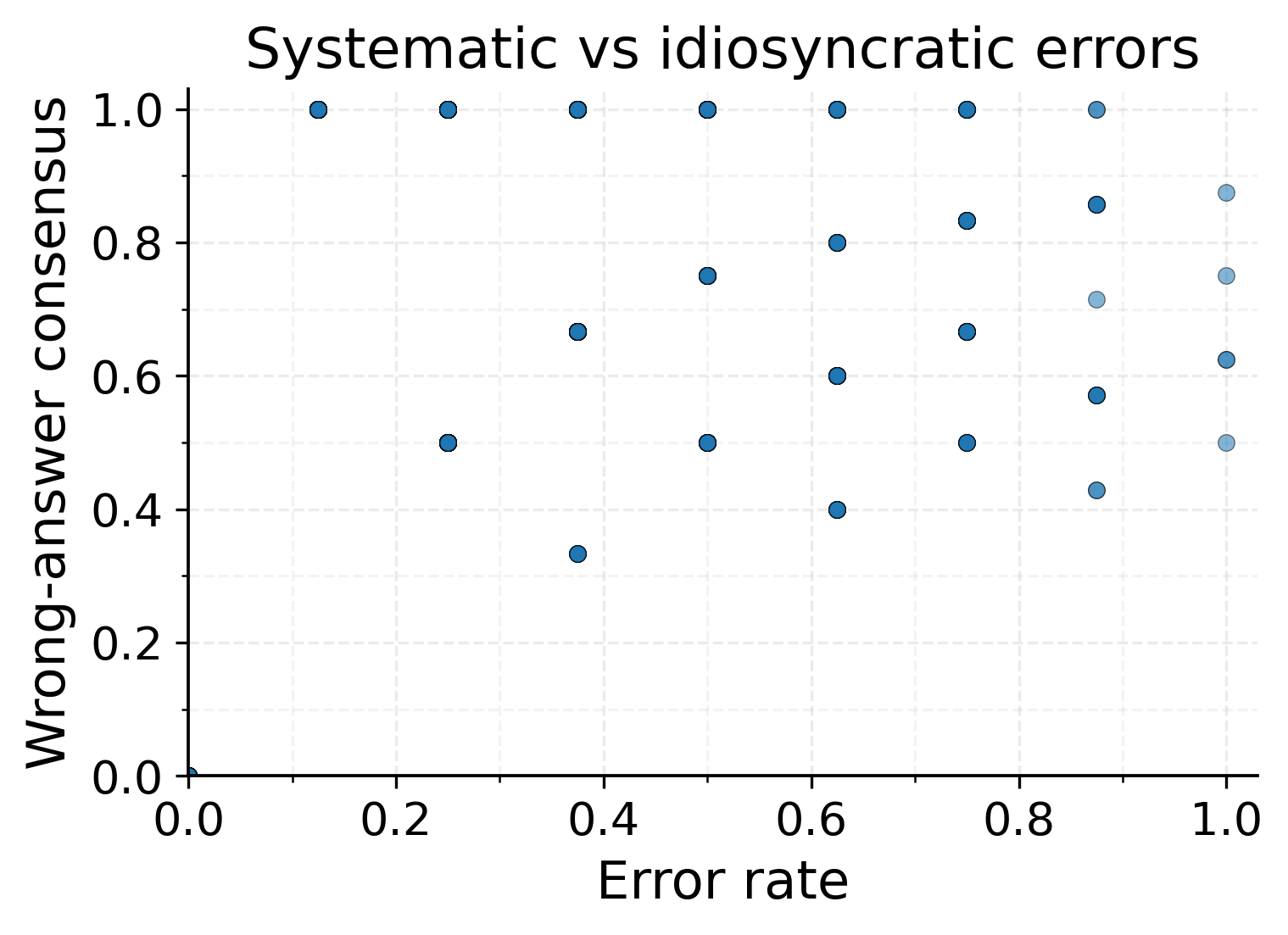}
        \caption{}
        \label{fig:error_vs_wrong_consensus}
    \end{subfigure}
    \hfill
    \begin{subfigure}[b]{0.49\linewidth}
        \centering
        \includegraphics[width=\textwidth]{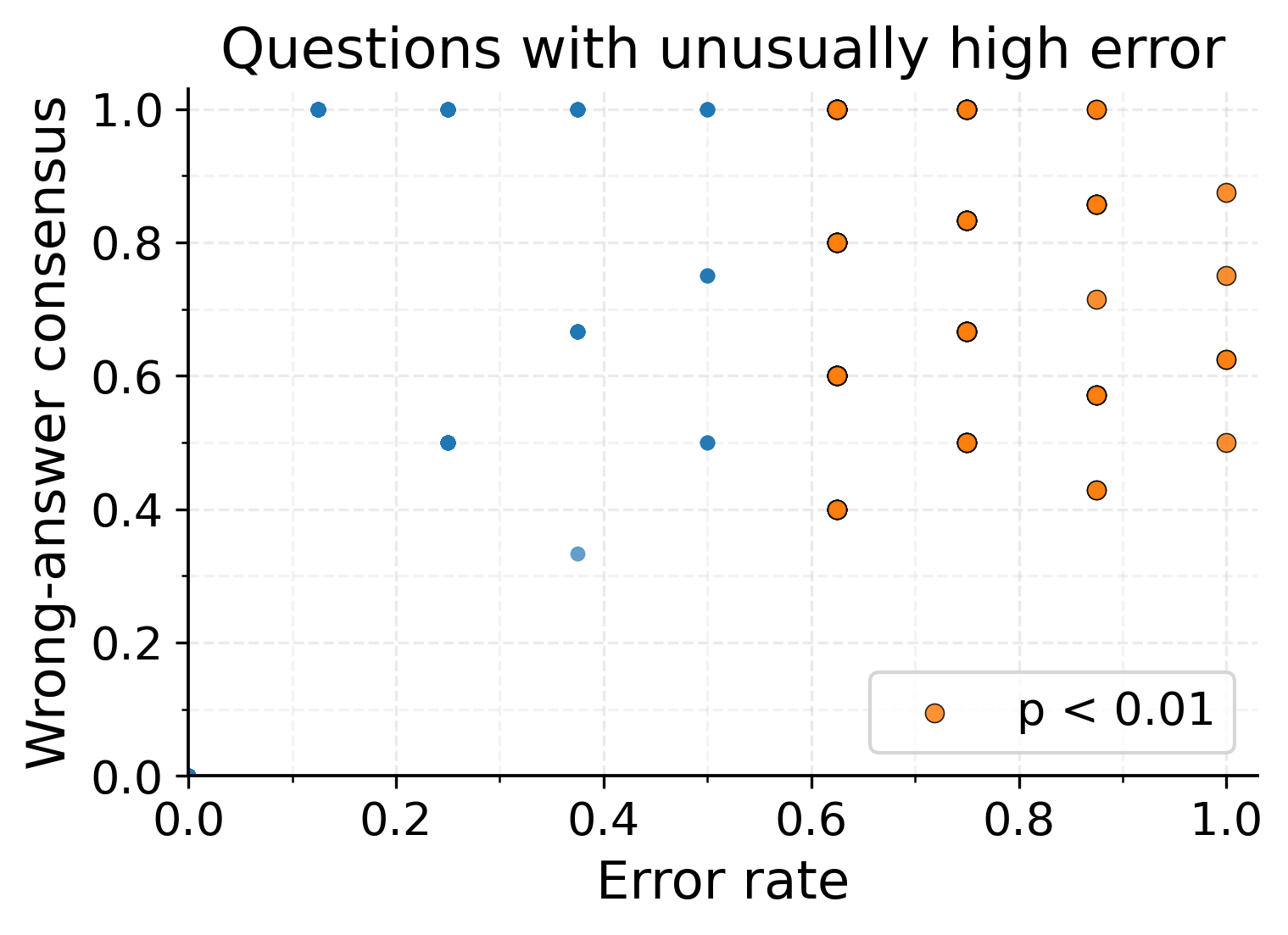}
        \caption{}
        \label{fig:error_vs_consensus_pvalue_highlight}
    \end{subfigure}}
    \caption{\textit{(a)} Per-question error rate versus wrong-answer consensus. Points in the upper-right region indicate questions for which multiple models fail in a consistent manner, signaling potential ambiguity or mislabeling. \textit{(b)} Error rate versus wrong-answer consensus with statistically anomalous questions highlighted (binomial test, $p<0.01$). The darker the spot, the more overlapping questions there are in that area.}
    \label{fig:error_vs_consensus}
\end{figure}

Figure~\ref{fig:error_vs_wrong_consensus} visualizes the relationship between per-question error rate and wrong-answer consensus (i.e. $\mathfrak{e}_i$ vs. $\mathfrak{c}_i$). Questions in the upper-right region of the plot correspond to systematic errors shared across models, and are therefore treated as primary candidates for manual review.

To further prioritize questions for inspection, we additionally compute a statistical anomaly score based on a binomial test under a null hypothesis in which each model answers correctly with probability equal to its empirical accuracy. Questions for which the observed number of incorrect predictions is highly unlikely under this null hypothesis are flagged as statistically anomalous. Figure~\ref{fig:error_vs_consensus_pvalue_highlight} highlights these questions, which represent the strongest candidates for dataset issues.

In the second stage, all questions flagged during automated screening are manually reviewed by the authors. Each flagged item is examined to verify the correctness of the mathematical content, the uniqueness of the correct answer, and the absence of ambiguity or underspecification. Questions found to contain errors or unclear formulations are either corrected or removed from the dataset.

After verification, the final dataset consists of $1{,}500$ validated questions. Each question is released together with its topic label, the complete set of answer options, and the index of the correct answer. To ensure that LLMs can correctly parse the mathematical content, and given the exposure of LLMs to \LaTeX{} data during training, all formulas are written in \LaTeX{}. The dataset is distributed in a machine-readable format suitable for large-scale evaluation, following the standard conventions of HuggingFace datasets. The full dataset is also available on GitHub in \texttt{json} format, where the questions are ready to be used for benchmarking and comparison with \texttt{lm\_eval}~\cite{eval-harness}.

\begin{table*}[t]
  \begin{center}
    \begin{small}
      \begin{sc}
        \begin{tabular}{lcccccc}
          \toprule
          \multirow{2}{*}{Model} & \multirow{2}{*}{Overall} & Linear & \multirow{2}{*}{Optimization} & Vector & \multirow{2}{*}{Probability} & \multirow{2}{*}{Python} \\
          & & Algebra & & Calculus & & \\
          \midrule
          Llama-3.1 8B Instruct & 70.4 & 64.0 & 71.6 & 67.0 & 82.5 & 62.1 \\
          Qwen2.5-Math 7B Instruct & 81.5 & 85.1 & 81.7 & 74.7 & 90.0 & 69.7 \\
          Qwen3 4B Instruct & 82.0 & 86.0 & 82.6 & 68.8 & 93.7 & 72.3 \\
          Qwen3 30B Instruct & 87.9 & 88.7 & 85.1 & \textbf{78.6} & 95.7 & 91.3 \\
          Qwen3-Coder 30B Instruct & \textbf{89.4} & \textbf{91.8} & \textbf{88.4} & 75.8 & \textbf{97.4} & \textbf{92.8} \\
          \midrule
          google-gemini-3-flash-preview & 86.5 & 82.6 & 83.2 & \textbf{83.5} & 92.6 & 92.3 \\
          openai-gpt-5 & 90.6 & 91.5 & \textbf{89.6 }& 77.9 & \textbf{99.1} & 94.4 \\
          anthropic\_claude-sonnet-4.5 & \textbf{90.9} & \textbf{93.9} & 84.1 & 80.7 & 98.3 & \textbf{99.0} \\
          \bottomrule
        \end{tabular}
      \end{sc}
    \end{small}
  \end{center}
  \caption{Normalized accuracy of various LLMs (both open- and closed-weights) over our dataset.}
  \label{tab:llms_accuracy}
  \vskip -0.1in
\end{table*}

\section{Experiment Settings}
\label{sec:experiments}
To evaluate our models on the curated dataset, we used the LM Evaluation Harness framework~\cite{eval-harness}. Given the multiple-choice nature of our benchmark, we adopted a zero-shot evaluation protocol based on log-likelihood ranking for open-weights models, while for closed-weights models, we relied on prompt-based text generation, since their logit probabilities are not accessible.

\subsection{Log-Likelihood Evaluation Mechanism}
For each multiple-choice question, evaluation is framed as a ranking problem over a fixed set of $N=3$ answer choices, rather than as a free-form generation task. Let $i$ indicate the question index, and denote by $x_i$ the input prompt (containing the question statement). Let $Y_i = \{y_i^{(1)}, \dots, y_i^{(N)}\}$ be the set of candidate answers, where each option $y_i^{(n)}$ corresponds to a sequence of $\ell_{i,n}$ tokens,
\begin{align*}
y_i^{(n)} = \bigl(y_{i,1}^{(n)}, y_{i,2}^{(n)}, \dots, y_{i,\ell_{i,n}}^{(n)}\bigr).
\end{align*}
For each option $y_i^{(n)} \in Y_i$, we compute its conditional log-likelihood under the model:
\begin{align*}
    \mathcal{L}\bigl(y_i^{(n)} | x_i\bigr)
    := \sum_{j=1}^{\ell_{i,n}} \log p\left(y_{i,j}^{(n)} \, | x_i, y_{i,1}^{(n)}, \dots, y_{i,j-1}^{(n)}\right).
\end{align*}
The predicted answer is then selected by maximizing the log-likelihood across the candidate options:
\begin{align}
\hat{y}_i = \arg\max_{y_i^{(n)} \in Y_i} \mathcal{L}\bigl(y_i^{(n)} | x_i\bigr).
\end{align}
In practice, we compute $\mathcal{L}\bigl(y_i^{(n)} | x_i\bigr)$ using teacher forcing, i.e., by scoring the target completion tokens one at a time conditioned on the prompt and the previously scored tokens.
We then report results using length-normalized log-likelihood scores to mitigate biases toward option length, defined as:
\begin{align}
\mathcal{L}_{\mathrm{norm}}\bigl(y_i^{(n)} | x_i\bigr)
:= \frac{1}{\ell_{i,n}} \mathcal{L}\bigl(y_i^{(n)} | x_i\bigr),
\end{align}
which prevents the model from being systematically biased toward shorter, high-probability completions (e.g., ``$(0,0)$'' vs.\ ``no stationary points'').

\subsection{Benefits of Log-Likelihood Ranking}\label{sec:benefits_loglikelihood}
Using log-likelihood ranking instead of standard text generation provides several advantages for objective multiple-choice benchmarking. First, generation-based evaluation can suffer from parsing failures: a model may produce the correct choice, but embed it in conversational text (e.g., ``The correct answer is (c)'') that a rigid post-processing rule may fail to extract. In contrast, log-likelihood ranking removes the need for answer parsing entirely, since evaluation is performed by directly scoring each candidate string. Second, generation-based protocols can be sensitive to decoding choices such as greedy decoding, nucleus sampling, or temperature, which introduce stochasticity and reduce reproducibility. Log-likelihood yields a deterministic score for every option, enabling consistent and repeatable evaluation. Third, the order and phrasing of answer options can disproportionately influence generation-based predictions, potentially leading to misleading results~\cite{raimondi2025exploiting}. Log-likelihood ranking mitigates this effect by explicitly comparing the model’s relative preference over all candidates.

\subsection{Prompt-Based Evaluation for Closed-Weights Models}
For closed-weights models, token-level probabilities (logits) are not accessible through the public interfaces we used. Therefore, unlike open-weights models, we evaluate closed-weights models via prompt-based answering, where the model is instructed to select one option among \texttt{(a)}, \texttt{(b)}, and \texttt{(c)} and to output only the chosen label.

In particular, for each question, we provide the full problem statement together with the three answer options, and we append an explicit constraint to enforce a machine-readable output format. Concretely, we instruct the model to reply with \emph{only} one of \texttt{(0)}, \texttt{(1)}, or \texttt{(2)} corresponding respectively to \texttt{(a)}, \texttt{(b)}, and \texttt{(c)} labels, constrained with an $<$\texttt{Answer}$>$ tag. If the output does not contain a valid number, or if the format is not respected, the prediction is considered invalid and marked as wrong. The resulting prompt fed to the model has the following structure:

\begin{quote}\small
\textbf{Question:} \emph{[question text]} \\
\textbf{Options:} \\
(0) \emph{[option A]} \\
(1) \emph{[option B]} \\
(2) \emph{[option C]} \\
\textbf{Instruction:} \\
Answer using only the following XML format:\\
$<$\texttt{Answer}$>$x$</$\texttt{Answer}$>$, where x is 0, 1, or 2.\\
If you produce any text outside the $<$\texttt{Answer}$>$ tag, the answer is invalid.\\
$<$\texttt{Answer}$>$
\end{quote}

To avoid variance due to sampling, we use deterministic decoding, setting the temperature to $\tau = 0.0$. All closed-weights evaluations are performed in a zero-shot setting.

\section{Results and Discussion}\label{sec:discussion}
The results provide a comprehensive overview of the current capabilities of LLMs across a diverse set of graduate-level mathematical domains. While a performance gap between open-weight models and proprietary state-of-the-art systems still persists, specialized open models are rapidly narrowing this gap in several categories.

Table~\ref{tab:llms_accuracy} reports the normalized accuracy of all evaluated models on \texttt{CompMath-MCQ}, both overall and across the five subject areas. Performance varies substantially across domains, highlighting differences in model strengths and weaknesses. We report length-normalized log-likelihood accuracy to reduce bias toward shorter answer strings, as described in Section~\ref{sec:experiments}. Overall, models achieve higher accuracy on \textit{Probability} and \textit{Python}, likely because these topics (including applied probability reasoning and programming-related content) are well represented in contemporary pre-training corpora. In contrast, \textit{Vector Calculus} and, to a lesser extent, \textit{Linear Algebra} remain more challenging, as they often require precise symbolic manipulation, careful bookkeeping across multiple variables, and multi-step reasoning.

\begin{figure*}[t]
    \centering
    \resizebox{\linewidth}{!}{
    \begin{subfigure}[b]{0.49\linewidth}
        \centering
        \includegraphics[width=\textwidth]{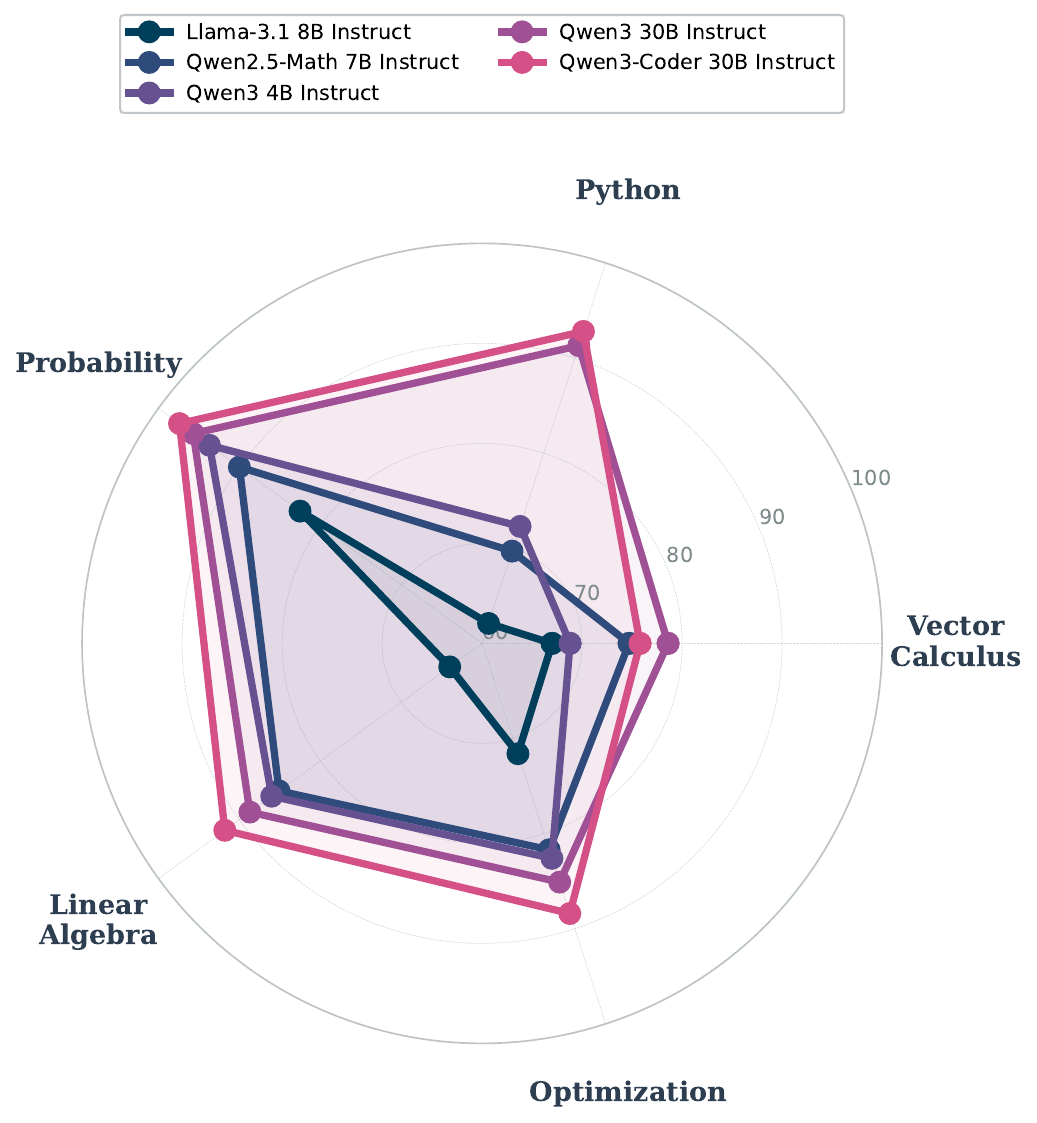}
        \caption{Open models accuracy.}
        \label{fig:radar_chart_open}
    \end{subfigure}
    \hfill
    \begin{subfigure}[b]{0.49\linewidth}
        \centering
        \includegraphics[width=\textwidth]{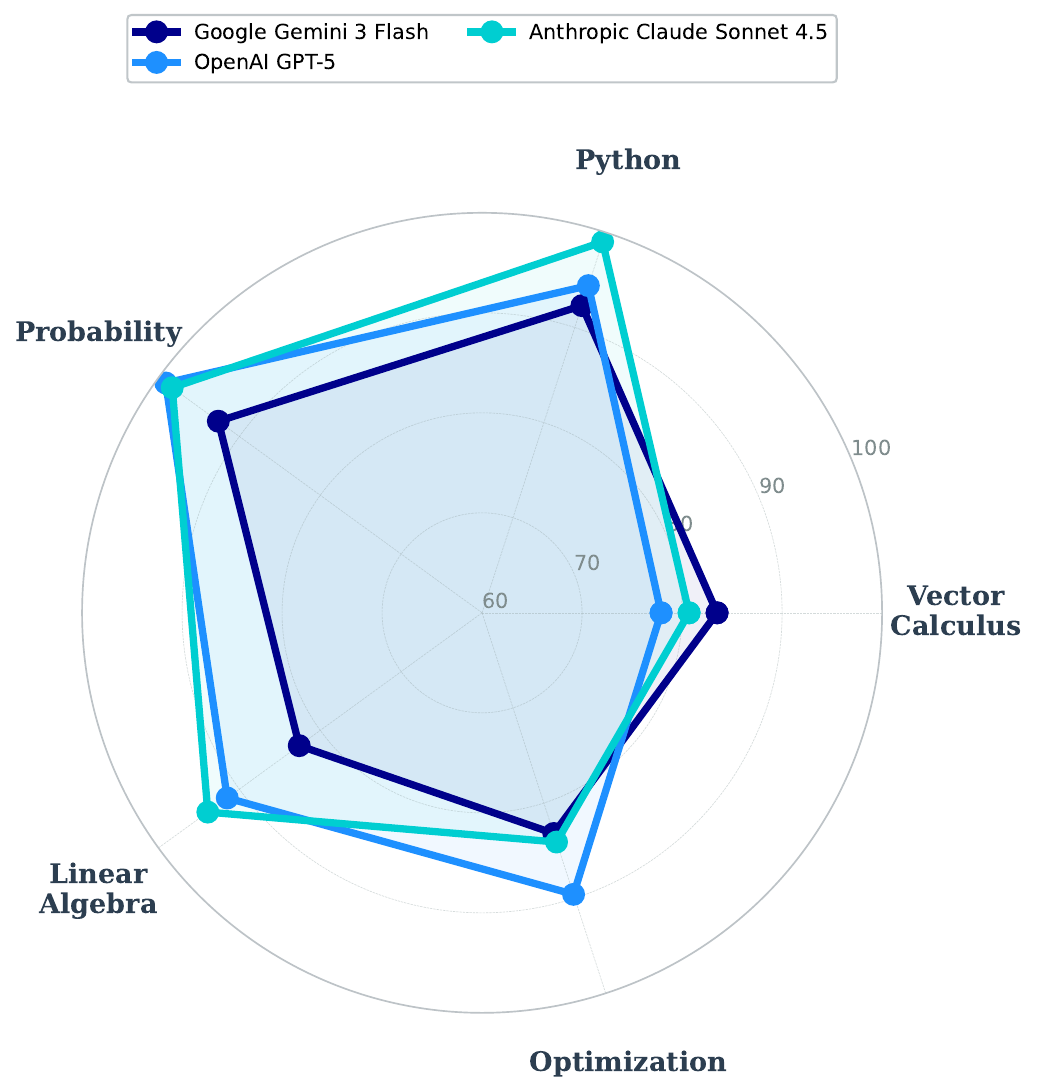}
        \caption{Closed models accuracy.}
        \label{fig:radar_chart_closed}
    \end{subfigure}}
    \caption{Comparison of LLMs accuracy across categories on \texttt{CompMath-MCQ}.}
    \label{fig:overall_radar_chart}
\end{figure*}

Figure~\ref{fig:overall_radar_chart} further illustrates cross-category variability. \textit{Probability} consistently yields the highest scores, with \texttt{GPT-5} and \texttt{Qwen3-Coder 30B Instruct} exceeding $99\%$ and $97\%$, respectively. This suggests that the combinatorial structures and reasoning patterns required in probability questions are relatively well captured by current training corpora and model representations.

Conversely, \textit{Vector Calculus} emerges as the most challenging category for most models. Even strong systems such as \texttt{Gemini 3 Flash} ($83.5\%$) and \texttt{Qwen3 30B Instruct} ($78.6\%$) perform noticeably worse in this domain than elsewhere. We hypothesize that this gap stems from the combination of multi-variable symbolic operations (e.g., gradients, Jacobians, and trigonometric compositions), spatial reasoning in higher-dimensional settings, and the need to carry out multiple intermediate steps reliably. By manually inspecting the answers given by the models, we additionally observe that many vector calculus errors correspond to sign mistakes, incorrect partial derivatives, or failure to correctly apply the chain rule, which are typical failure modes of LLMs in multi-step symbolic reasoning.

Among proprietary models, \texttt{GPT-5} and \texttt{Claude Sonnet 4.5} show strong overall performance ($90.6\%$ and $90.9\%$, respectively), with relatively stable accuracy across categories. This consistency may indicate more robust internal representations of core mathematical concepts and improved reliability in step-wise computations compared to smaller open-weight models.

Among open-weight models, \texttt{Qwen3-Coder 30B Instruct} achieves the best overall accuracy ($89.4\%$), approaching the performance of proprietary systems despite being fully reproducible. We also observe that domain-specialized models (e.g., \texttt{Qwen2.5-Math 7B}) can outperform larger general instruction models (e.g., \texttt{Llama-3.1 8B}) across most categories, highlighting the impact of math-focused post-training. Interestingly, \textit{Optimization} appears less challenging than \textit{Vector Calculus}, suggesting that gradient-based reasoning and stationary-point concepts are more consistently handled than multi-variable symbolic differentiation.

Finally, the inclusion of \textit{Python} questions highlights the increasingly intertwined nature of mathematical reasoning and scientific computing. The strong performance of code-oriented models such as \texttt{Qwen3-Coder 30B Instruct} ($92.8\%$ in \textit{Python} and $89.4\%$ overall) suggests that programming proficiency correlates with stronger performance on computational reasoning tasks, supporting the ``code-as-thought'' hypothesis \cite{yang2025code}. Interestingly, \texttt{Claude Sonnet 4.5} achieves the best \textit{Python} accuracy among the closed-weights models ($99.0\%$), which may contribute to its strong results in \textit{Linear Algebra} by enabling more effective internal simulation of matrix operations and numerical routines.

\section{Conclusion and Future Work}\label{sec:conclusion}
We introduced \texttt{CompMath-MCQ}, a leakage-free, curriculum-aligned benchmark of $1{,}500$ originally authored multiple-choice questions targeting graduate- and PhD-level \emph{computational} mathematics. To improve reliability, we proposed a two-stage validation pipeline combining cross-model agreement statistics with manual expert review. For evaluation, we used \texttt{lm\_eval} with normalized log-likelihood ranking for open-weight models and format-constrained prompt-based option selection for closed-weight models whose token-level probabilities are not accessible. Baseline results reveal substantial domain differences: LLMs performance is strongest on \textit{Probability} and \textit{Python}, while \textit{Vector Calculus} (and parts of \textit{Linear Algebra}) remains a bottleneck, reflecting persistent difficulties with multi-step symbolic manipulation and careful bookkeeping.

\bibliography{biblio}
\bibliographystyle{icml2026}

\end{document}